\pdfoutput=1

\documentclass[11pt]{article}

\usepackage[]{acl}

\usepackage{times}
\usepackage{latexsym}

\usepackage[T1]{fontenc}

\usepackage[utf8]{inputenc}
\usepackage[caption=false]{subfig}

\usepackage{microtype}
\usepackage{lipsum}  

\usepackage{inconsolata}

\usepackage{xcolor}
\usepackage{array,multirow,graphicx}
\usepackage{float}
\usepackage{soul}

\usepackage{enumitem}

\newcommand{\GG}[1]{}

\title{Asking the Right Question at the Right Time: Human and Model Uncertainty Guidance to Ask Clarification Questions}

\author{Alberto Testoni \and Raquel Fern\'{a}ndez \\
  Institute for Logic, Language and Computation \\
  University of Amsterdam \\
  \texttt{\{a.testoni, raquel.fernandez\}@uva.nl} \\}

\begin{document}
\maketitle
\begin{abstract}

Clarification questions are an essential dialogue tool to signal misunderstanding, ambiguities, and under-specification in language use. While humans are able to resolve uncertainty by asking questions since childhood, modern dialogue systems struggle to generate effective questions. To make progress in this direction, in this work we take a collaborative dialogue task as a testbed and study how model uncertainty relates to human uncertainty---an as yet under-explored problem. We show that model uncertainty does not mirror human clarification-seeking behavior, which suggests that using human clarification questions as supervision for deciding when to ask may not be the most effective way to resolve model uncertainty. To address this issue, we propose an approach to generating clarification questions based on model uncertainty estimation, compare it to several alternatives, and show that it leads to significant improvements in terms of task success. Our findings highlight the importance of equipping dialogue systems with the ability to assess their own uncertainty and exploit in interaction. 

\end{abstract}

\section{Introduction}

\begin{figure*}[!ht]
\centering
  \includegraphics[width=1\linewidth]{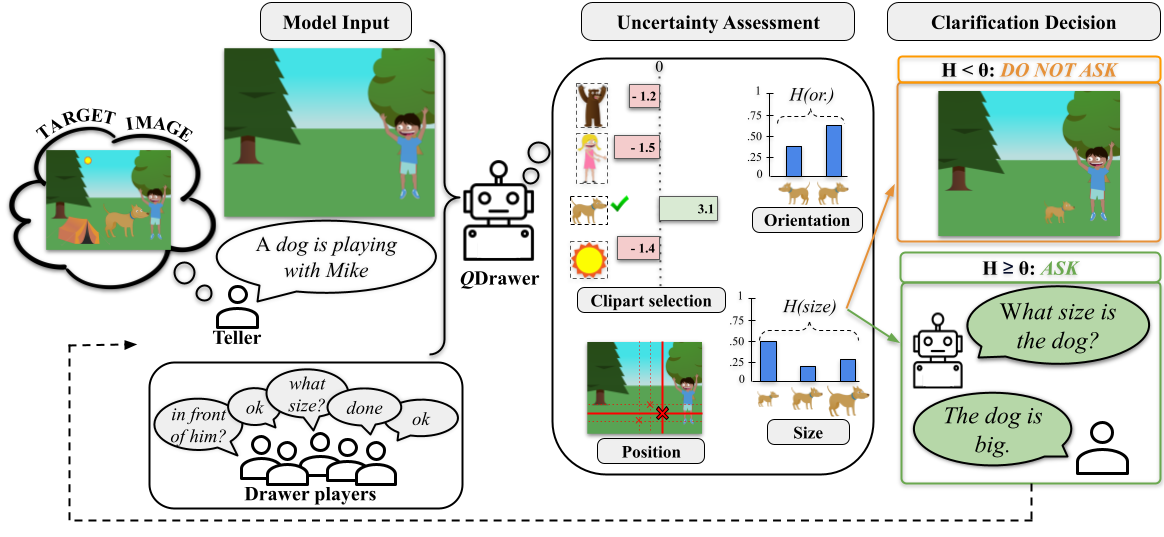}
  \caption{Overview of our experimental setup in CoDraw. After receiving an instruction from the Teller, the Drawer agent selects the clipart(s) to draw, together with their attributes. If the entropy over an attribute exceeds a threshold $\theta$ (\textit{size} in the figure), then the Drawer asks a clarification question. The question-answer pair is added to the dialogue history before performing the next drawing action. Different human players may react differently (bottom-left box); the agent decides whether to ask for clarification on the basis of its own uncertainty, independently from the clarification decisions of human players.}
  \label{fig:intro_img}
\end{figure*}

The ability to ask effective and informative questions is a peculiar feature of human intelligence. From an evolutionary perspective, it has been argued that being able to ask questions to reliable informants gives humans a key evolutionary advantage over other species \citep{tomasello2009cultural}. Questions are also shown to play a fundamental role in cognitive development and language learning during childhood \citep{demetras1986feedback, chouinard2007children, leech2013father, ruggeri2016sources, ruggeri2017toma}. 
Both children and adults generally ask questions to resolve a state of uncertainty \citep{piaget1954construction, rothe2018people}. 
Deciding how to resolve uncertainties and maintaining mutual understanding are essential steps to achieve \textit{conversational grounding} \citep{clark1987collaborating, clark1991grounding}, the process through which humans negotiate the production and acceptance of their utterances. Clarification questions are among the key strategies speakers use to achieve conversational grounding. However, while humans handle and resolve uncertainties through clarification questions almost effortlessly in everyday conversations, this is particularly challenging for modern dialogue systems based on deep neural networks \citep{chandu-etal-2021-grounding}. Even the new generation of conversational agents, such as ChatGPT,\footnote{\url{https://chat.openai.com/}} seem to struggle to generate clarification questions when dealing with ambiguous instructions \citep{deng2023prompting}.

In this paper, we investigate uncertainty-driven strategies for asking clarification questions. Clarification requests are usually triggered when ambiguity and/or underspecification make it difficult for the receiver to follow up with enough confidence \citep{purver2004theory,benotti2017modeling, pezzelle2023semantic}. It should come as no surprise then that there is a long-standing tradition of work highlighting the importance of incorporating clarification requests in dialogue systems \cite[][among others]{paek2000conversation, gabsdil2003clarification, schlangen2004causes, rieser2005implications, rieser-lemon-2006-using}. 
Recent years have witnessed a renewed interest in this ambitious goal, especially in the context of collaborative dialogue tasks \cite[e.g.,][]{nguyen-daume-iii-2019-help, chi2020just, zhu2021self, shi-etal-2022-learning, shen2023learning, madureira-schlangen-2023-instruction}. Yet, several key aspects remain under-explored. 

For example, some existing approaches use a dedicated neural network model trained on human clarification behaviour to decide \textit{when} to ask \citep{shi-etal-2022-learning,madureira-schlangen-2023-instruction}. However, it is well known that humans may follow different confirmation-seeking strategies \citep{schegloff1977preference, schegloff1987some, purver-etal-2001-means}. This potential human disagreement casts doubt on the effectiveness of using human decisions as training signal for supervised or reinforcement learning. As an alternative approach, model uncertainty has often been employed to decide \textit{when} and/or \textit{what} to ask \citep{nguyen-daume-iii-2019-help, zhu2021self, shen2023learning, chi2020just}. Yet the two types of approaches have not been directly compared, nor is there any study (to the best of our knowledge) on the extent to which model uncertainty correlates with human uncertainty. In our work, we take a closer look at these fundamental aspects. 

We consider the CoDraw task and dataset \citep{kim-etal-2019-codraw}---where a Drawer player is asked to recreate a clipart scene following instructions from a Teller player---as a testbed to study uncertainty-driven clarification strategies in collaborative agents. We propose different ways to represent model uncertainty in CoDraw, and then carry out a study on the relation between model uncertainty and human uncertainty, as measured by the decision to ask clarification requests in human-human dialogues. 
This study reveals that there is a poor relation between model and human uncertainty, which in part appears to stem from our finding that in CoDraw human players do not follow a consistent clarification strategy. 
This poor relation suggests that using human clarification decisions as supervision for deciding when to ask may not be the most effective way to resolve model uncertainty. 
We therefore propose a new Drawer agent model equipped with an uncertainty-based clarification module (illustrated in Figure~\ref{fig:intro_img}), where we use a template-based approach to generate clarification questions triggered by the uncertainty state of the agent. This approach allows us to modulate the level of uncertainty that leads to posing a question and thus control for the number of questions asked per dialogue, which may be advantageous given the processing cost of questions for humans as well as models \citep{chiyah-garcia2023sigdial}. 

Our results and analyses show that our proposed model is significantly more effective in terms of task success than several strong baselines, including a version based on the human supervision model by \citet{madureira-schlangen-2023-instruction}. Overall, our findings highlight the importance of equipping conversational agents with the ability to assess and act upon their own uncertainty by asking effective clarification questions. 

\section{Related Work}
\label{sec:related}





Clarification requests have long been studied as a key repair mechanism in human dialogue \citep{gabsdil2003clarification, purver2004theory, schlangen2004causes, rieser-lemon-2006-using, schloder2014clarification, benotti2017modeling, benotti-blackburn-2021-recipe}.
Datasets of human conversations where spontaneous clarification questions are annotated have been an important resource for data-driven research on clarification strategies. For example, \citet{rodriguez2004form} proposed an annotation scheme based on Clark's four levels of communication \cite{clark1996using} to identify the form and function of clarification requests. 
These and similar annotation schemes have been applied to 
small datasets \citep{benotti2009clarification, gervits-etal-2021-agents, benotti-blackburn-2021-recipe}. 

At a larger scale, the TEACh dataset \citep{padmakumar2022teach} has been annotated by \citet{gella-etal-2022-dialog} with dialogue acts, including acts related to clarification, and the Minecraft Dialogue Corpus \citep{narayan-chen-etal-2019-collaborative} and its extensions \citep{shi-etal-2022-learning, kiseleva2022iglu, mohanty2022collecting} have been annotated with different types of clarification requests. However, in the latter case the dataset was augmented with further clarification questions \textit{a-posteriori}, introduced asynchronously by other annotators who were asked to decide whether clarifications were needed. \citet{aliannejadi-etal-2021-building} and \citet{gao2022dialfred} adopt similar techniques for including clarification questions in dialogue data.

Recently, \citet{madureira-schlangen-2023-instruction,madureira2023you} released CoDraw-iCR (v1 and v2), a fine-grained annotation of spontaneous clarification requests in the CoDraw dataset \citep{kim-etal-2019-codraw}. This constitutes a worthwhile resource for analysing clarification behaviour in collaborative tasks. \citet{madureira-schlangen-2023-instruction} exploit these annotations to train a neural model for deciding \textit{when} to clarify. In contrast, we shall use them to study the relation between human and model uncertainty in Section~\ref{sec:human} and then, in 
Section~\ref{sec:asking_clarification_questions}, propose a model for the CoDraw Drawer agent that asks clarification questions on the basis of its own uncertainty.

Model uncertainty represents a valuable signal for Natural Language Generation models to deal with the different sources of ambiguity in language use \citep{baan2023uncertainty}. Earlier on, when the speech recogniser used to be a common source of errors in spoken dialogue systems, ASR confidence scores 
were exploited to decide on error handling strategies such as confirmation and clarification \cite[][among others]{skantze2008galatea,stoyanchev2014towards}. 
In more recent systems, where the focus is on the next action to take in a collaborative task, a common approach consists of using model uncertainty (as measured by the entropy over the model prediction on a set of actions) to signal ambiguity and trigger the generation of clarification questions \citep{nguyen-daume-iii-2019-help, zhu2021self, shen2023learning}. Without explicitly estimating uncertainty, \citet{testoni-bernardi-2021-looking} propose a beam search re-ranking algorithm guided by the model intermediate predictions about the target in a referential task. Similar approaches use the probability difference between the top two predicted actions to estimate uncertainty \citep{chi2020just}. Recently, \mbox{\citet{naszadi-etal-2023-aligning}} investigated the alignment between predictive uncertainty and ambiguous instructions in visually-grounded communication tasks. They show that well-calibrated prediction probabilities benefit the detection of ambiguous
instructions. Differently from our work, the authors do not investigate the effectiveness of model uncertainty to generate follow-up clarification questions.

Instead of relying on model uncertainty, other proposed methods involve training external neural networks or sub-modules with the objective of deciding \textit{when} to ask for clarification \cite{madureira-schlangen-2023-instruction, shi-etal-2022-learning}. \citet{khalid-etal-2020-combining} propose to incorporate cognitive modeling in a Reinforcement Learning framework to generate context-sensitive clarification strategies in a referential game. While a few approaches generate clarification questions from scratch in an end-to-end fashion \cite{khalid-etal-2020-combining, zhu2021self}, heuristics are often applied at the generation stage \citep{shen2023learning, zhang2021diverse, sekulic2021towards}. 
In this paper, we propose an approach to generating clarification questions based on model uncertainty estimation and compare it to alternative methods for deciding when to clarify by training on human clarification decisions.

\section{Task \& Dataset} 
\label{sec:data}


CoDraw \citep{kim-etal-2019-codraw} is a goal-driven instruction-following collaborative task between two players. The Teller player has access to an abstract target scene containing a variable number of cliparts (on average, target scenes contain 6 cliparts), leveraging the Abstract Scene dataset \citep{zitnick2013bringing, zitnick2013learning}. The Teller provides written instructions in English to the Drawer in order to reconstruct the scene in a turn-based fashion. The Drawer has access to a gallery of 58 cliparts that they can place on a canvas. The Drawer can remove, resize (small, medium, large), or flip (facing right or left) each object in the scene. After receiving an instruction, the Drawer has to reply to the Teller before proceeding to the following turn. Typically, the Drawer either replies with confirmation feedback or asks follow-up questions. The CoDraw dataset consists of around 10k dialogues, with an average length of 7.7 turns per dialogue. Additional details can be found in Appendix \ref{app:codraw-data-models}.

In CoDraw, task success is measured by comparing the target image with the one drawn by the Drawer at the end of the conversation. To this aim, \citet{kim-etal-2019-codraw} designed a similarity score metric, ranging from 0 to 5, tailored to the CoDraw task. This score takes into account different cliparts (by checking whether cliparts in the target scene appear in the reconstructed one) and their attributes (size, position, flipping, plus some additional attributes for cliparts representing people, i.e., facial expression and body pose) by assigning different weights to each component. 
The scenes drawn by human players reach a similarity score of 4.17 (sd=0.64). In our work, we apply some changes to the similarity score metric to fix some flaws in its design and better capture the properties under analysis in the experimental setup, as described in Appendix \ref{appendix:ss}. In the following, we report the results using the new version of the metric.


\section{Computing Uncertainty in CoDraw}
\label{sec:uncertainty}

\subsection{The Silent Drawer Model} 
\label{sec:silent}

\citet{kim-etal-2019-codraw} proposed a neural Drawer model for the CoDraw task.
At each round of the dialogue, this model is conditioned on the Teller’s latest message, which is encoded into a vector using a bidirectional LSTM module. The Drawer also receives as input a vector representation of the current state of the canvas that is being drawn (the cliparts added so far and their attributes). These input representations are passed to a dense feed-forward neural network. The resulting vector represents the Drawer’s follow-up action, which determines the cliparts to be added to the canvas, together with their attributes. Thus, \citeauthor{kim-etal-2019-codraw}'s Drawer model is \textit{silent}, i.e., 
it can only draw objects on the canvas but cannot contribute to the dialogue. 
The model is trained in a Supervised Learning fashion on human data. At inference time, the model assigns an unbounded score to each of the 58 available cliparts: each clipart with a score above 0 is drawn on the canvas, together with its predicted attributes (size, position, orientation, etc.). The model is tested by providing turn-by-turn Teller's instructions from the CoDraw test set, reaching an average similarity score of 3.31 (sd=0.67).

\subsection{Computing Model Uncertainty}
\label{sec:uncertainty}

We are interested in representing the uncertainty of the model regarding its next action. After receiving an instruction, the Drawer model has to make decisions about which clipart(s) to draw and their corresponding attributes. For each dialogue turn and selected cliparts, we estimate model uncertainty regarding these components as follows:

\begin{itemize}[leftmargin=11pt,itemsep=1pt]
    \item For \textbf{clipart selection}, we take the score assigned to each clipart 
    as a proxy of model uncertainty. The lower the value, the more uncertain 
    the model is about selecting the given clipart. 
    \item For the \textbf{size} (\textit{small, medium, large}) and \textbf{orientation} (\textit{right, left}) attributes, we compute the entropy of the probability distribution over the possible labels per attribute. The higher the entropy, the higher the uncertainty on that attribute. 
    \item For the \textbf{position} attribute, the Drawer outputs two values per clipart, corresponding to the $x$- and $y$-axis coordinates of the canvas. Measuring uncertainty from continuous values is inherently more challenging than discrete variables. 
    We take inspiration from uncertainty estimation using Ensembles of different models \citep{lakshminarayanan2017simple} and consider the variance among the prediction of five Drawer models trained with different random initialization seeds as a representation of model uncertainty. 
\end{itemize}
\section{Model vs.~Human Uncertainty} 
\label{sec:human}


In this section, we investigate the relation between model and human uncertainty. In the absence of clues such as eye gaze or reaction times (which could be indicative of uncertainty, but are not available in CoDraw), we take clarification questions as a signal of human uncertainty. We exploit the annotation of clarification questions released by \citet{madureira-schlangen-2023-instruction, madureira2023you}, which consists of turns by the Drawer player annotated as clarification requests plus the attributes mentioned in such requests (size, position, orientation, etc.).\footnote{In particular, we take advantage of the annotation in CoDraw-iCR (v2) by \citet{madureira2023you}.} Around 40\% of the dialogues in CoDraw contain at least one clarification question, with an average number of 2.2 clarification requests per dialogue in this subset.

\subsection{Setup}
\label{subsec:uncertainty_setup}
We explore whether we can use model uncertainty (as defined in the previous section) to predict when human players ask clarification questions. To this aim, 
we train a logistic regression model with human decisions about whether to ask a clarification question in the follow-up turn as \textit{dependent variable} and each model uncertainty type (clipart selection, size, orientation, position) as \textit{independent variables}. We use the Drawer model described in Section \ref{sec:silent} to compute model uncertainty at each dialogue turn. 
Using the methods described in Section~\ref{sec:uncertainty}, we compute uncertainty for each clipart with a score above 0 and each of its attributes. If this includes more than one clipart at a given turn, we take the highest uncertainty value per variable (clipart selection, size, orientation, position).\footnote{Other approaches, such as taking the average of the uncertainty values, do not lead to any significant difference in the final results.}
We use the CoDraw test set (1002 dialogues, with an average of 7.7 turns per dialogue) to extract model uncertainty. We further split the CoDraw test set to train and evaluate the logistic regression model (70\% and 30\%, respectively). 
We study the effect of different initialization seeds and training epochs and compare the model performance to a random baseline.

\begin{figure}
  \includegraphics[width=\linewidth]{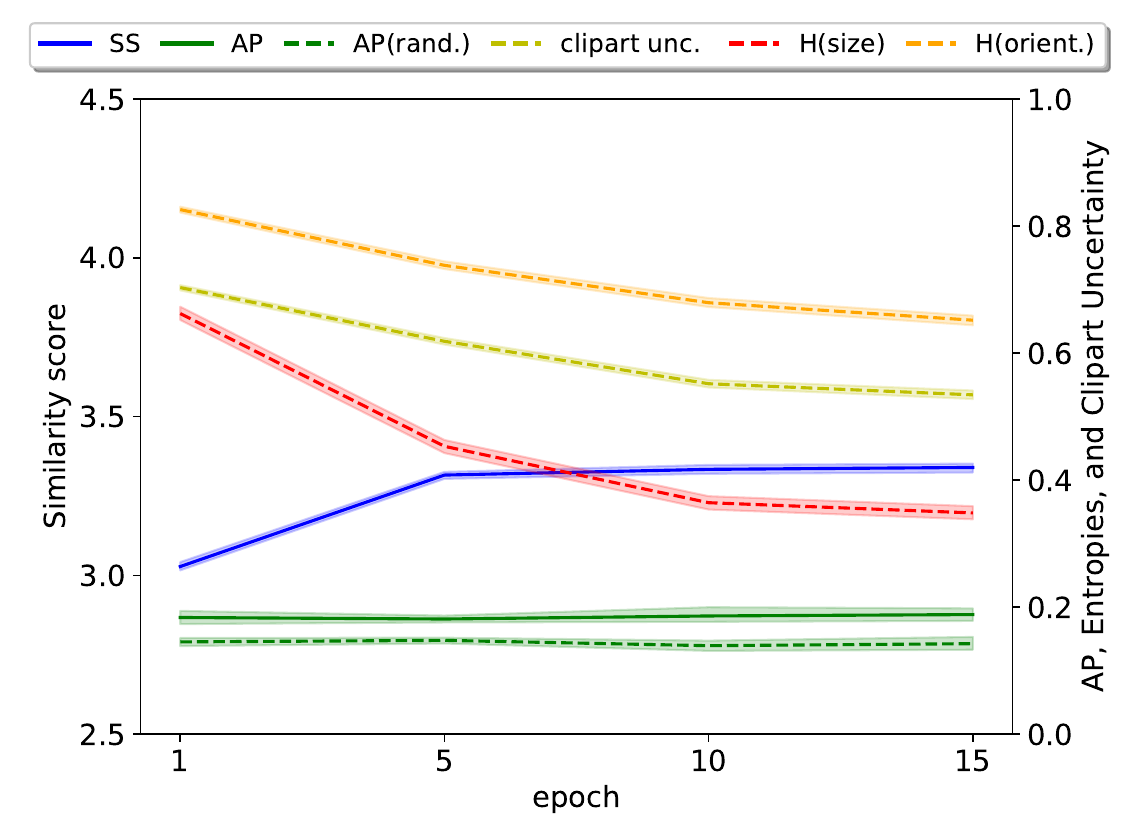}
  \caption{
  Average Precision (\textit{AP}, green) of a logistic regression model in predicting when human players asked clarification questions based on the uncertainty extracted from a Drawer trained for a variable number of epochs ($x$-axis). 
  The plot also shows the similarity score (\textit{SS}, blue) achieved by the Drawer, and its uncertainty on clipart selection (normalized and reversed clipart scores), and on the size and orientation attributes (as measured by the entropy \textit{H}). For each epoch, we report the average and variance of 5 Drawer models trained with different seeds. Note that the $y$-axis has two scales: we report SS on the left, while all the other metrics refer to the right side.}
  \label{fig:regression_res}
\end{figure}

\subsection{Results} 
\label{subsec:uncertainty_results}

The results are displayed in Figure~\ref{fig:regression_res}. 
Let us consider the solid and dotted green lines in the plot. The logistic regression model trained to predict \textit{when} human players ask clarification questions based on the uncertainty of the Drawer model yields an Average Precision of 0.188 (epoch 15; average across 5 different initialization seeds). This result is above the random baseline (0.188 vs.~0.142, $p < 0.05$), indicating that model uncertainty has some degree of predictive power. Nevertheless, performance is poor. Among the independent variables analyzed, only the uncertainty about clipart selection has statistically significant predictive power ($p < 0.001$). Additional details about the coefficients can be found in Appendix~\ref{appendix:model_vs_human_uncertainty}.


Figure~\ref{fig:regression_res} provides additional interesting insights: we can see that the Drawer model learns to perform the task over the training epochs, as evidenced by the upward trend of the similarity score, which measures task success. At the same time, the uncertainty of the model decreases (the plot shows uncertainty on clipart selection, size, and orientation. Same pattern is observed for position). Thus, the Drawer model is reasonably well calibrated: higher task performance goes hand in hand with lower uncertainty. Yet, this trend does not have an impact on the power of model uncertainty to predict human clarification behaviour, as the  Average Precision of the logistic regression model remains constantly low over training epochs. We verified that other methods to capture model uncertainty, such as extracting epistemic and aleatoric uncertainty \citep{kendall2017uncertainties} using an Ensemble of models, lead to the same results and do not have significant predictive power. Additional details can be found in Appendix \ref{appendix:model_vs_human_uncertainty}.

\subsection{Analysis}
\label{subsec:uncertainty_analysis}
There could be many reasons why model uncertainty and human uncertainty are poorly related. Human players do not only have immensely more knowledge and resources than the relatively simple Drawer model; different speakers may also follow different conversation strategies and such variation may be difficult to predict. It is well known that there is variability in human clarification-seeking and repair strategies \citep{schegloff1977preference, schegloff1987some, purver-etal-2001-means}.
To check whether such variation is present 
in the CoDraw dataset, we carry out an exploratory analysis. We use a keyword-matching approach to cluster together instructions by Teller players with similar form and content that lack some key information, and check how frequently they are followed by clarification questions. We only consider Teller's instructions uttered in the first dialogue turn to exclude the possible confounding effect of the previous dialogue history. 
Full details on the procedure and the results can be found in Appendix \ref{appendix:model_vs_human_uncertainty}. Here we highlight the main takeaways. For example, we observe that there are a total of 2274 Teller first instructions that mention at least one clipart but do not specify its size (e.g., \emph{`a tree to the left a few inches from the edge'}). Only 23\% of follow-up turns by the Drawer player include a clarification question, and among these, only 27\% are about the size of the clipart. Similarly, there are 240 Teller first instructions that mention at least one clipart but do not specify its location (e.g., \emph{`little girl running towards her friend that has a pie in his hand'}). 49\% of follow-up turns include a clarification question, and among these, 48\% are about the location of the clipart.

Thus we observe that when analyzing utterances that do \textit{not} mention a specific necessary attribute, the human Drawer players may or may not ask about the missing information. Humans may be more or less cautious, take more or fewer risks, or guess/hope that more information will be provided later on in the dialogue \citep{purver2004theory, benotti2017modeling, benotti-blackburn-2021-recipe}.
This high human disagreement on deciding whether to clarify or not suggests that learning when to ask clarification questions from human clarification behaviour may not be optimal. Indeed the model proposed by \citet{madureira-schlangen-2023-instruction} achieves only 0.35 Average Precision.
Instead, in the next section we investigate whether model uncertainty represents an effective guidance to decide \textit{when} to ask clarification questions by studying their impact on task success.

\section{Asking Clarification Questions} 
\label{sec:asking_clarification_questions}

\begin{figure}
\centering
  \includegraphics[width=0.8\linewidth]{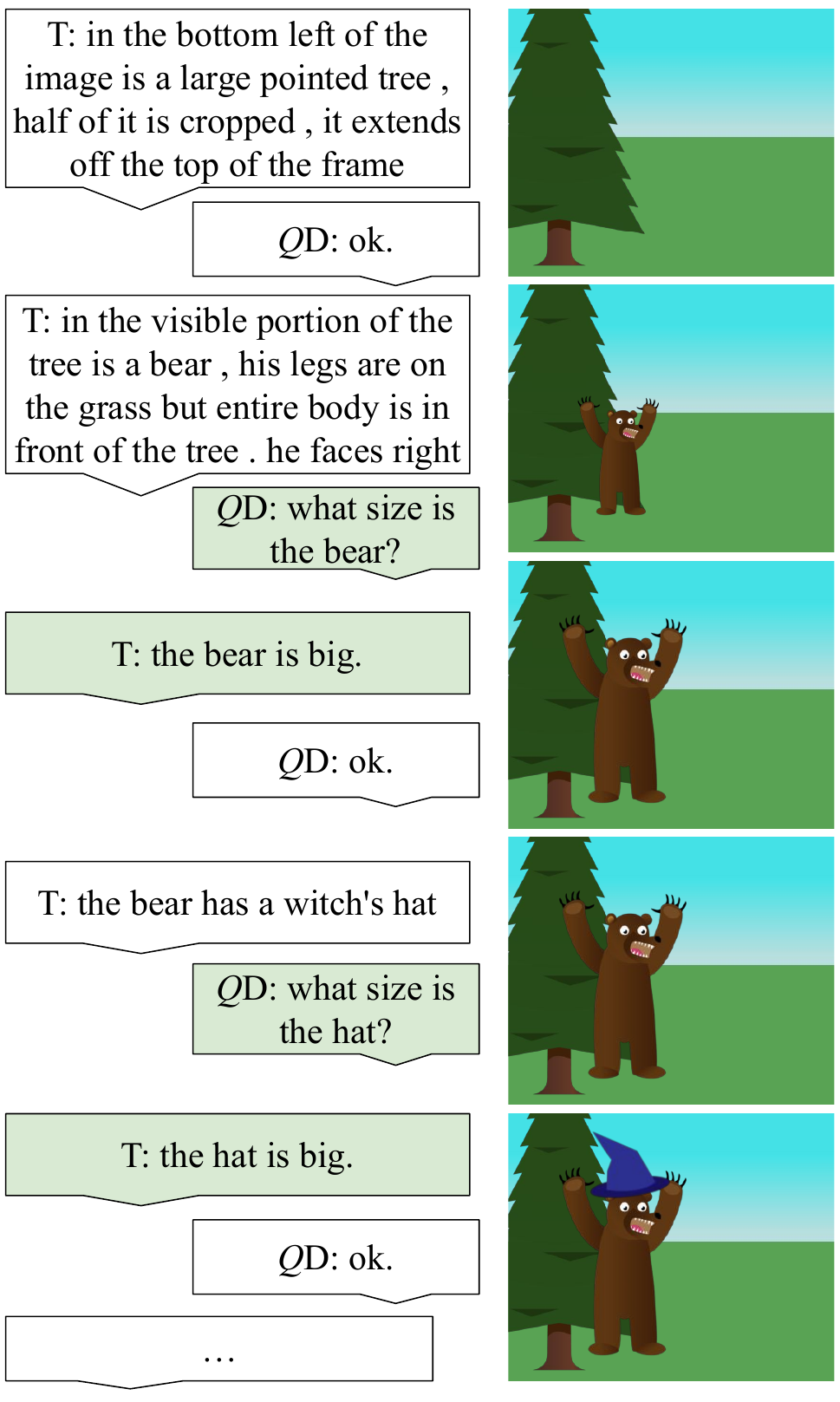}
  \caption{Example of clarification exchanges (in green) with questions generated by \textit{Q}Drawer and answers by the Teller. The images on the right depict the \textit{Q}Drawer’s canvas after each round of conversation.}
  \label{fig:dialogue_example_small}
\end{figure}

We implement a \textit{Questioning} Drawer model (\textit{Q}Drawer) by adapting the \textit{Silent} Drawer described in Section \ref{sec:silent}. We focus on the size attribute, because it applies to all cliparts and it is often underspecified in the Teller's instructions (see Section~\ref{sec:human}). Our goal is to test whether asking clarification questions triggered by size uncertainty results in (1)~higher accuracy regarding the size attribute, and (2)~higher overall similarity score. To isolate the effect of asking questions and avoid confounds related to language generation, we use a template-based approach to generate clarification questions and their answers.

\subsection{The Uncertainty-based \textit{Q}Drawer} 
\label{sec:qdrawer}

We adapt the \textit{Silent} Drawer as follows: (a) the model receives as input the immediately preceding turn exchange between the Drawer and Teller, instead of just the Teller's utterance; (b) we add a term in the cross-entropy loss to include not only the cliparts that are added to the canvas but also the ones that are updated by changing some attributes;\footnote{In this way, the model learns to assign a high score to all cliparts mentioned in the instruction and not just to the ones that do not appear yet on the canvas---an essential step to successfully profit from clarification exchanges. See Appendix \ref{appendix:QDrawer} for additional details.}
(c) we include a clarification generation module that works as follows:  
At any turn of the conversation, if the entropy over the size attribute labels (\textit{small}, \textit{medium}, or \textit{large}) exceeds a given threshold,\footnote{We experiment with several thresholds, as it will become clear in subsequent sections.} \textit{Q}Drawer generates a clarification question targeting a maximum number of 2 cliparts (i.e., the ones with the highest entropy). 

We employ the following template, which corresponds to the most common formulation of size questions in CoDraw according to \citet{madureira2023you}: \textit{``what size is the \_\_?''} or \textit{``what size are the \_\_ and the \_\_?''}, where the slots are filled with the name(s) of the clipart(s).  
We exploit the ground-truth annotation in the CoDraw dataset to provide an answer to the question on behalf of the Teller using these templates: \textit{``The \_\_ is \_\_''} or \textit{``The \_\_ is \_\_ and the \_\_ is \_\_''}, where the slots are filled with the name(s) of the clipart(s) under discussion and their size (small, medium, large). Before performing a new drawing action,  \emph{Q}Drawer is conditioned on the current state of the canvas, as well as 
the clarification question asked and the corresponding answer. Figure {\ref{fig:dialogue_example_small}} shows an example of clarification questions asked by \textit{Q}Drawer and the answers received, as well as the updated canvas after each round of the conversation. Another example is reported in Appendix \ref{appendix:asking_cq}.

We experimented with replacing the Bi-LSTM language encoder of the Drawer agent (see Section~\ref{sec:silent}) with more complex models (both fine-tuned on CoDraw and trained from scratch), such as BERT and RoBERTa \citep{devlin-etal-2019-bert, liu2019roberta} and their corresponding distilled versions \citep{sanh2019distilbert}. We found that these BERT-based \textit{Q}Drawer variants give rise to the same patterns on the role of clarification discussed in the results below, but lead to a deterioration of model performance as measured by the similarity score (max.~3.07).\footnote{This result is in line with a strand of work highlighting that LSTM-based models outperform BERT-like ones for small datasets and/or specific domains  \citep{zeyer2019comparison, ezen2020comparison, le2021hierarchical, wei2022visual}.} 
We thus decided to keep the original model architecture.\footnote{
Again, a logistic regression model trained with the uncertain extracted from the Drawer variants based on BERT and RoBERTa poorly correlates with human behaviour on asking clarification questions, reaching an average precision (0.18 AP) in line or slightly below the one discussed in Section \ref{sec:human}.} 


\subsection{Baselines}
\label{sec:baselines}

We compare \emph{Q}Drawer to the \textit{Silent} Drawer model, as well as to different configurations of \emph{Q}Drawer that differ with respect to the information used to decide \textit{when} to ask a question. 
In particular, we compare our proposed uncertainty-based approach to the following alternatives: 

\begin{itemize}[leftmargin=11pt,itemsep=0pt,topsep=2pt]
\item The supervised learning approach by \citet{madureira-schlangen-2023-instruction}, where an external neural model trained on human clarification behaviour is used to decide when to ask (\textit{Decider}). In this case, size uncertainty is still used to select which clipart(s) to target in the question, and the clarification question is generated and answered in the same way as in the uncertainty-based \emph{Q}Drawer. 

\item Asking questions at the same position (in terms of dialogue turn) where \textit{human} players ask questions about size in CoDraw. 

\item Asking questions at \textit{random} dialogue turns (discussed in Section \ref{subsec:how_many}).


\end{itemize}

\begin{table}[]
\centering
\resizebox{1\columnwidth}{!}{
\begin{tabular}{l|c|c|c|c|c|}
\cline{2-6}
                              & \textbf{\begin{tabular}[c]{@{}c@{}}When\\ To Ask\end{tabular}}        & \textbf{\begin{tabular}[c]{@{}c@{}}Size\\ Acc.\end{tabular}} & \textbf{\begin{tabular}[c]{@{}c@{}}SS\end{tabular}} & \textbf{\begin{tabular}[c]{@{}c@{}} CQs Size \\ Acc. boost\end{tabular}}& \textbf{\begin{tabular}[c]{@{}c@{}}CQs\\ SS boost\end{tabular}} \\ \hline
\multicolumn{1}{|l|}{\emph{S}D} & --                  & 76.9  & 3.31 & -- & --        \\ \hline
\multicolumn{1}{|l|}{\multirow{2}{*}{\emph{Q}D}} & Human              & 77.3 & \ \ 3.31$^{\circ}$  & \ +6.2\%   & \ \ +.04$^{\bullet}$       \\ \cline{2-6} 
\multicolumn{1}{|l|}{} & Decider            & 80.1 & 3.34  & \ +7.1\% & +.06         \\ \hline \hline
\multicolumn{1}{|l|}{\multirow{3}{*}{\emph{Q}D}} & $\theta = 0.3$ & 87.3 & 3.40  & +11.0\%   & +.09        \\ \cline{2-6} 
\multicolumn{1}{|l|}{} & $\theta = 0.7$ & 86.0 & 3.39  & +11.0\% & +.09          \\ \cline{2-6} 
\multicolumn{1}{|l|}{} & $\theta = 1.1$ & 82.4 & 3.36  & \ +9.8\% & +.09           \\ \hline
\end{tabular} 
}
\caption{Silent (\emph{S}D) and Questioning Drawer (\emph{Q}D) results using different approaches to generate clarification questions. Top part: baselines. \textit{CQs}: clarification questions. \textit{Human}: dialogue turns when human players ask CQs. \textit{SS}: similarity score on the whole test set. \textit{Size acc.}: accuracy (\%) of the size attribute of all drawn cliparts on the whole test set. \textit{Boost} columns: SS and Size Acc. boost brought by CQs on the \textit{subset} of dialogues containing at least one CQ. All the differences are statistically significant except $^{\circ}$ (non-significant difference with SD) and $^{\bullet}$ (adding CQs does not significantly increase SS on the subset).}
\label{tab:results}
\end{table}

\subsection{Task Success Results}
\label{subsec:size_acc}

Table \ref{tab:results} summarizes the results of our experiments on the CoDraw test set. We compare different models on their accuracy in correctly selecting the size of the cliparts drawn on the canvas and the overall similarity score between the drawn and the target scene at the end of the dialogue. 
The \textit{Silent} Drawer (\textit{S}D) reaches a size accuracy of 76.9\% 
and a similarity score of 3.31. The version of \emph{Q}Drawer that generates size questions in the same position as human players 
only marginally improves performance on size accuracy, while the similarity score does not show any statistically significant increase (paired t-test, $p > 0.05$). Using an external neural \textit{Decider} to choose \textit{when} to ask clarification questions improves the performance over \textit{S}D (up to 80.1\% accuracy and 3.34 similarity score). Finally, the uncertainty-driven \emph{Q}Drawer outperforms the baselines to a large extent: up to 87.3\% size accuracy and 3.40 similarity score.\footnote{ 
An exploratory study on extending our approach to other attributes reveals that using both \textit{size} and \textit{orientation} uncertainty further increases the similarity score on the test set up to 3.46, indicating that our proposed method can scale up.} 
Regarding the comparison with the \textit{Decider} model, it is worth noting that this model is trained to ask questions at each position where human players asked questions, not only when humans asked size questions. While this method is somewhat coarse-grained (because the \textit{Decider} may ask size-CQ when they are not needed), it ensures full recall (the \textit{Decider} will ask a size question whenever humans asked a size question). Hence, the \textit{Decider} baseline is still meaningful with respect to the size attribute.

The \emph{Q}Drawer model shows a consistent improvement as the entropy threshold is lowered. This aspect is investigated in more detail in Section~\ref{subsec:how_many}.

\subsection{Clarification Questions' Contribution}

To further analyze how clarification exchanges contribute to task success, for each model configuration, we consider the subset of dialogues ${CQ}$ where at least one clarification question is asked. We then remove all clarification questions from these dialogues, resulting in $!{CQ}$. We compute the difference in size accuracy and similarity score between ${CQ}$ and $!{CQ}$ to check the specific contribution of clarification questions (accuracy and SS boost). As we can see on Table \ref{tab:results}, the boost brought about by clarification questions is much higher for the uncertainty-driven \emph{Q}Drawer than for the baseline models. In particular, for the subset of dialogues in  ${CQ}$, clarification exchanges increase size accuracy by $+11\%$ in absolute terms and the similarity score by $+0.09$. Asking questions in the same position as human players does not significantly increase the similarity score when comparing ${CQ}$ and $!{CQ}$ (paired t-test, $p>0.05$). 
To make sure that this statically non-significant result does not stem from our generation approach, we replace machine-generated clarification questions with human ones for the dialogues in ${CQ}$. We find that in this case the increase in both size accuracy and similarity score is even less pronounced than what we observe in Table~\ref{tab:results}, resulting in non-significant results for both question-generation methods.

We further investigate the impact of clarification exchanges by analyzing the calibration of the model uncertainty estimates. We compare the probability distribution over the size attribute against the ground-truth size of each object by using two metrics: the Expected Calibration Error \cite[ECE,][]{naeini2015obtaining, guo2017calibration} and the Brier score \mbox{\citep{brier1950verification}}. For both metrics, a lower score indicates better calibration.\footnote{We refer to \mbox{\citet{ovadia2019can}} for a broader discussion on calibration metrics.} 
We compare the \textit{Silent} Drawer and \emph{Q}Drawer ($\theta = 0.3$) and find that the \textit{Silent} Drawer reaches an ECE value of 0.10 while 
the ECE for \emph{Q}Drawer goes down to 0.05 (the difference is significant: t-test, $p<0.001$). 
Similarly, the Brier score significantly improves (t-test, $p<0.001$) from 0.36 (\textit{Silent} Drawer) to 0.28 (\emph{Q}Drawer). These results indicate that asking clarification questions improves the calibration of the model uncertainty estimates.

\begin{table}[]
\centering
\resizebox{0.9\columnwidth}{!}{
\begin{tabular}{|c|c|c|}
\hline
\textbf{$\theta$} & \textbf{\begin{tabular}[c]{@{}c@{}}\% dialogues with\\ at least 1 CQ\end{tabular}} & \textbf{\begin{tabular}[c]{@{}c@{}}avg. number of \\ CQ per dialogue\end{tabular}} \\ \hline
0.3   & 96.6 & 3.63           \\\hline
0.7   & 84.7 & 2.69              \\\hline
1.1   & 57.1 & 1.84               \\\hline                                                                                         
\end{tabular}
}
\caption{The effect of different $\theta$ values on the number of questions asked by \emph{Q}Drawer. We report the percentage of dialogues with at least one clarification question and, within this subset, the average number of clarification questions per dialogue.}
\label{tab:number_of_q}
\end{table}

\begin{figure}[!t]
    \centering
  \includegraphics[width=1\linewidth]{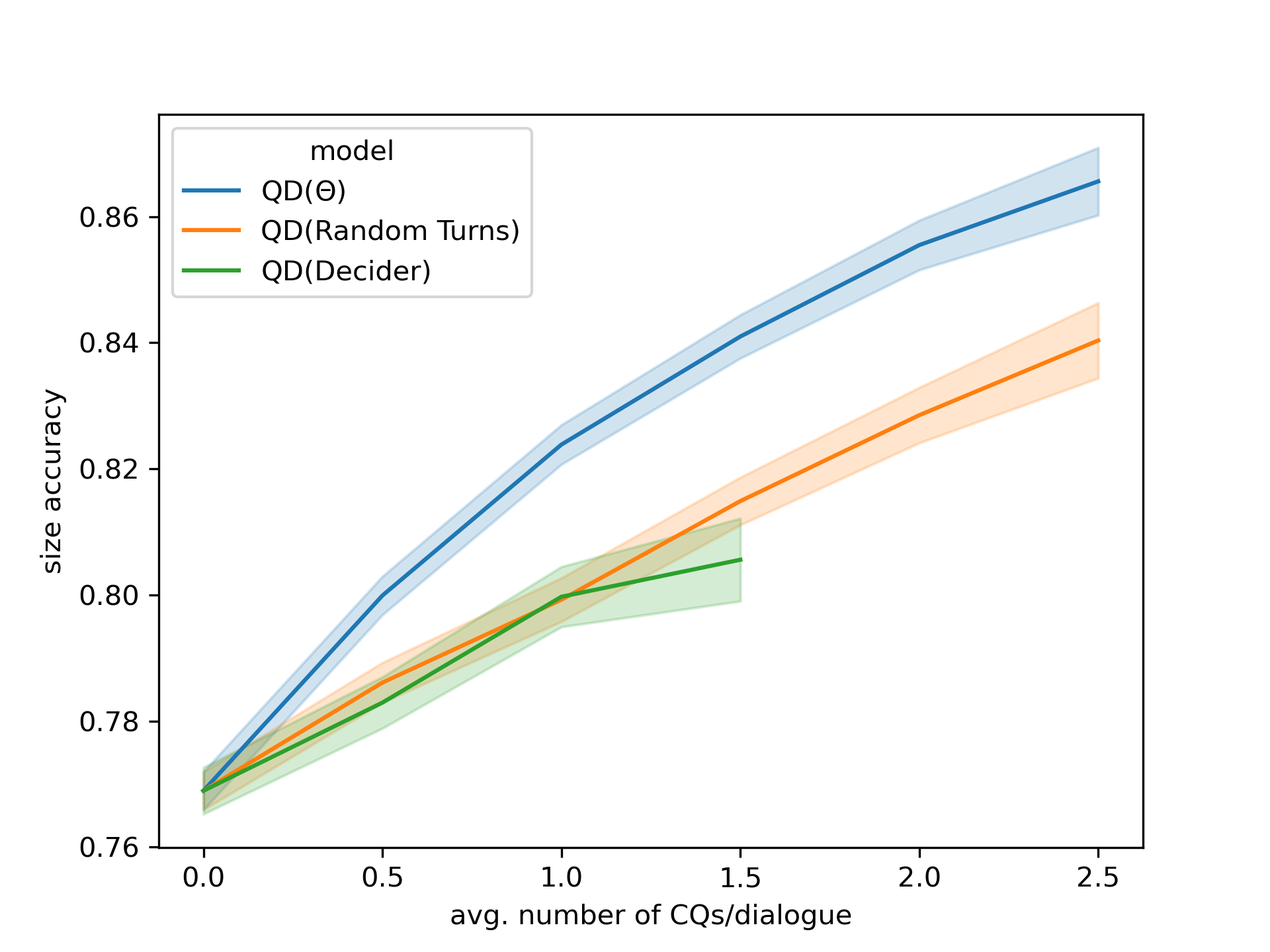}
  \caption{Effect of the average number of questions per dialogue (considering all dialogues in the test set) on the \textit{size} accuracy. We compare the uncertainty-guided \emph{Q}Drawer with a version that asks questions in random turns and one that asks questions in the turns selected by an external \textit{Decider} model.}
  \label{fig:num_qs_effect}
\end{figure}

\subsection{\textit{How Many} Questions to Ask} 
\label{subsec:how_many}

Asking clarification questions carries a cost, both from the perspective of the agent asking the question and the agent processing it \citep{clark1996using, purver2004theory}.
We take this partially into account by adopting a simple approach: we evaluate \emph{Q}Drawer by using different entropy threshold values $\theta$ to control for the \textit{number} of questions generated by the model. As reported in Table \ref{tab:number_of_q}, in \emph{Q}Drawer the value of the entropy threshold $\theta$ has, by design, a direct impact on the number of questions asked: i.e., a model that only asks for clarification in the face of high uncertainty ($\theta=1.1$) gives rise to fewer dialogues with questions and to a lower average number of questions per dialogue. 

We compare the uncertainty-driven \emph{Q}Drawer against (1)~a version that asks questions in random dialogue turns with different average number of questions per dialogue and (2) the \emph{Q}Drawer paired with the \textit{Decider} module as described in Section~\ref{sec:baselines}. For the latter, the maximum possible number of questions per dialogue (average of 1.5) is defined by the \textit{Decider} model output. Figure \ref{fig:num_qs_effect} shows the effect of the number of questions on size accuracy for different model configurations. While the performance of the \textit{Decider} model almost overlaps with the \emph{Q}Drawer asking questions in random dialogue turns, using model uncertainty significantly improves accuracy, even when a few questions are asked. This result confirms that even when controlling for the number of questions asked (which may be desirable in terms of cost control), the uncertainty-driven \emph{Q}Drawer shows the most promising results.

\section{Conclusions}
\label{sec:conclusions}


Different kinds of uncertainties characterize language use, and asking informative and effective questions is a crucial tool to address these uncertainties. While humans are able to ask strategic questions since early childhood to resolve a state of uncertainty, equipping modern dialogue systems with the ability to ask effective questions has proved challenging. Some existing approaches propose employing dedicated neural network models trained on human behaviour to decide when to ask for clarification in a dialogue. However, the high variability in human clarification strategies casts doubt on the effectiveness of this approach. Other  proposed methods  involve using model uncertainty to decide \textit{when} and/or \textit{what} to ask. Yet, estimating and exploiting model uncertainty is not trivial, and the relationship between model and human uncertainty is still unclear. 

In our work, we considered the CoDraw collaborative dialogue task as a test bed. We proposed different ways of extracting model uncertainty and investigated its relationship to human uncertainty, as measured by the clarification decisions of human players. We found that human and model uncertainty are poorly correlated, possibly due to the wide diversity of human clarification strategies. We then presented an approach that makes use of model uncertainty to generate clarification questions, and compared it to other methods for deciding when clarification is needed. Our approach allows us to vary the level of uncertainty that elicits the generation of clarification questions to control for the number of questions generated. The results of our experiments show that our proposal outperforms a number of baselines and other methods trained on human clarification decisions. More generally, our findings pave the way for further research on integrating model uncertainty into natural language generation approaches as an effective way to resolve ambiguities and address underspecification in language use \citep{pezzelle-2023-dealing}.

\section*{Limitations}
\label{sec:limitations}

Even though our experimental setup (which focuses on the size attribute using a template-based generation approach) allows us to isolate the effect of clarification exchanges in a controlled setting, we acknowledge that further work is required to confirm our findings in more complex collaborative tasks and model architectures. In our work, we focused on questions about the \textit{size} attribute, and we mentioned some results about expanding our approach to the \textit{orientation} attribute to assess the robustness of our approach. Assessing the contribution of clarification exchanges is particularly challenging given the wide range of factors involved, both on the appropriateness of the question asked and the correctness of the answer received. Without some restrictions, it is difficult to disentangle the role of these factors and this may threaten a fair model evaluation. For this reason, we focus on the \textit{size} attribute, as it applies to all cliparts and it is often underspecified in the Teller's instructions (see Section~\ref{sec:human}). An additional advantage of focusing on the size attribute is that we can ensure the correctness of the answer (see the templates in Section~\ref{sec:qdrawer}), while for other attributes such as position there is a less direct mapping between natural language expressions and space coordinates. 

Regarding the generalization of our setup to other settings, we would like to highlight that, although simple, the CoDraw task shares many features with a wide variety of different tasks and datasets. In particular, in Vision-and-Language Navigation tasks an agent has to reach a goal by moving around an environment and performing specific tasks. We believe this setting, where the agent has a set of actions at their disposal, shares many similarities with the CoDraw setup. On the other hand, in more complex tasks the role of the visual input may play a greater role compared to CoDraw, leading to new challenges in evaluating the grounding of the generated question-answer exchanges. At the same time, we acknowledge the fact that a template-based approach may very well not scale up to other settings. However, in our work, this choice is motivated by the focus on the effect of clarifying rather than the question generation problem.

Finally, we only used a simple approach to control for the ``cost'' of asking clarification questions by limiting the overall number of possible questions. We acknowledge that a more fine-grained approach is required to consider the wide variety of different factors that influence the cost of generating clarification requests. Given the simplicity of the CoDraw task and the structure of the questions our method generates, we do not analyze how clarification questions may influence the perceived naturalness of the generated dialogues. We acknowledge that this aspect should be taken into account when dealing with more complex tasks and generative strategies.

\section*{Acknowledgments}
We would like to thank Brielen Madureira for discussions about the CoDraw task and dataset, as well as for sharing insights on CoDraw-iCR. We are grateful to Patrick Kahardipraja for sharing an updated version of the original CoDraw codebase. Finally, we thank the Dialogue Modelling Group (DMG) at the University of Amsterdam for their feedback and support at the different stages of this work. This project has received funding from the European Research Council (ERC) under the European Union’s Horizon 2020 research and innovation programme (grant agreement No.~819455).
\bibliography{anthology,custom}

\appendix

\section*{Appendix}
\label{sec:appendixA}

\section{Similarity Score (v2)}
\label{appendix:ss}
In our work, we apply some changes to the similarity score metric proposed in \citet{kim-etal-2019-codraw} to fix some flaws in its design and better capture the properties under analysis in the experimental setup. More specifically, (a) we assign the same weights to all components of the similarity score; (b) we consider the \textit{orientation} attribute only for those cliparts that are not perfectly symmetrical; (c) we change the denominator of the \textit{facial expression} and \textit{body pose} attributes so that they apply only to cliparts representing people. Using the new metric, scenes drawn by human players reach a similarity score of 4.42 (sd=0.62).

\section{Model vs.~Human Uncertainty}
\label{appendix:model_vs_human_uncertainty}
\paragraph{Logistic Regression results and analysis} Table \ref{table:regression_ap_ss} shows the Average Precision of the logistic regression model trained to predict when human players ask clarification questions in CoDraw based on model uncertainty. We extract uncertainty from Drawer models trained for different training epochs and initialization seeds. We also report the performance of a random baseline and the CoDraw similarity score for each configuration. Similarly, Table \ref{tab:f1} reports the F1 score of the same model. Table \ref{tab:coeff} shows the correlation coefficients of the logistic regression model for each attribute-specific uncertainty analyzed (clipart selection, size, orientation, position). We report the coefficients for each model configuration (initialization seed + training epoch). Finally, Table \ref{tab:clusters} illustrates some insights and examples of the ``clusters'' analyzed in Section \ref{sec:human} to investigate the variability of human confirmation-seeking strategies.

\paragraph{Epistemic and Aleatoric uncertainty} We also experimented with including Epistemic and Aleatoric uncertainty in the Logistic Regression analysis. We obtain the same results as using the other uncertainty measures described above, without any statistically significant predictive power. Similarly to \citet{greco-etal-2022-small}, we use an Ensemble of models trained with different initialization seeds to compute \textit{data} (aleatoric) and \textit{model} (epistemic) uncertainty \citep{depeweg2018decomposition, hullermeier2021aleatoric} for the size and orientation attributes. The ensemble \emph{total uncertainty} is computed as the entropy of its predictive distribution $[p_{\theta_n}(y | x)]$, which averages the distributions of the single ensemble components:
\begin{equation}
H[p(y|x)] = H[1/N \sum_{n=1}^N p_{\theta_n}(y | x)].
\end{equation}
\emph{Data uncertainty} is measured as the average uncertainty of each ensemble component:
$1/N \sum_{n=1}^N H [p_{\theta_n}(y | x)]$. The difference between total uncertainty and data uncertainty results in \emph{model uncertainty}. As mentioned above, including these uncertainties in the logistic regression model described in Section \ref{sec:human} does not improve its Average Precision, which remains significantly above chance level but low in absolute values (0.19 AP, similar to the values reported in Figure \ref{fig:regression_res}).
 
\section{Asking Clarification Questions}
\label{appendix:asking_cq}
\begin{figure*}[t]
  \centering
  \subfloat{
  \includegraphics[width=0.98\linewidth]{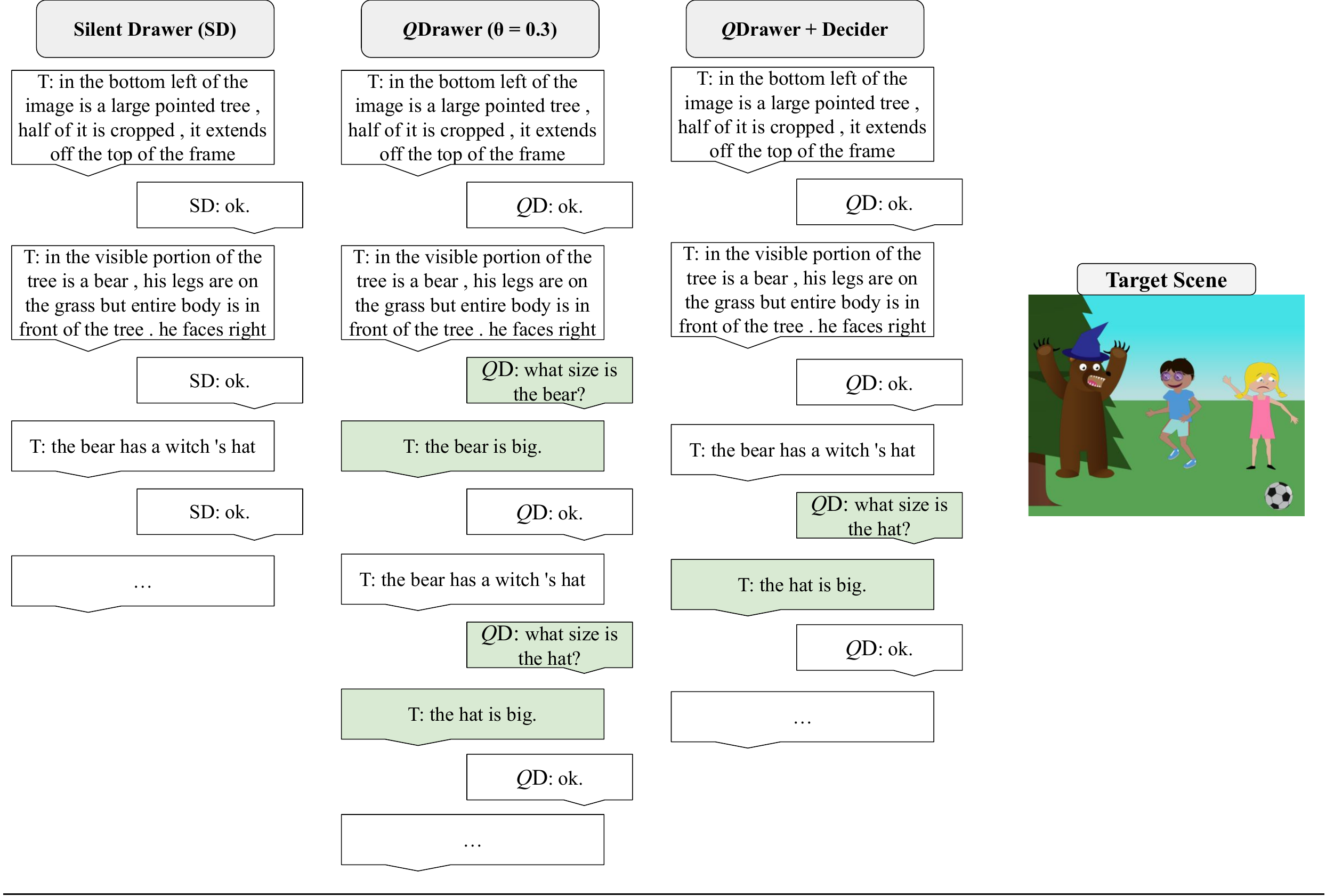} }\\
  \subfloat{
  \includegraphics[width=0.98\linewidth]{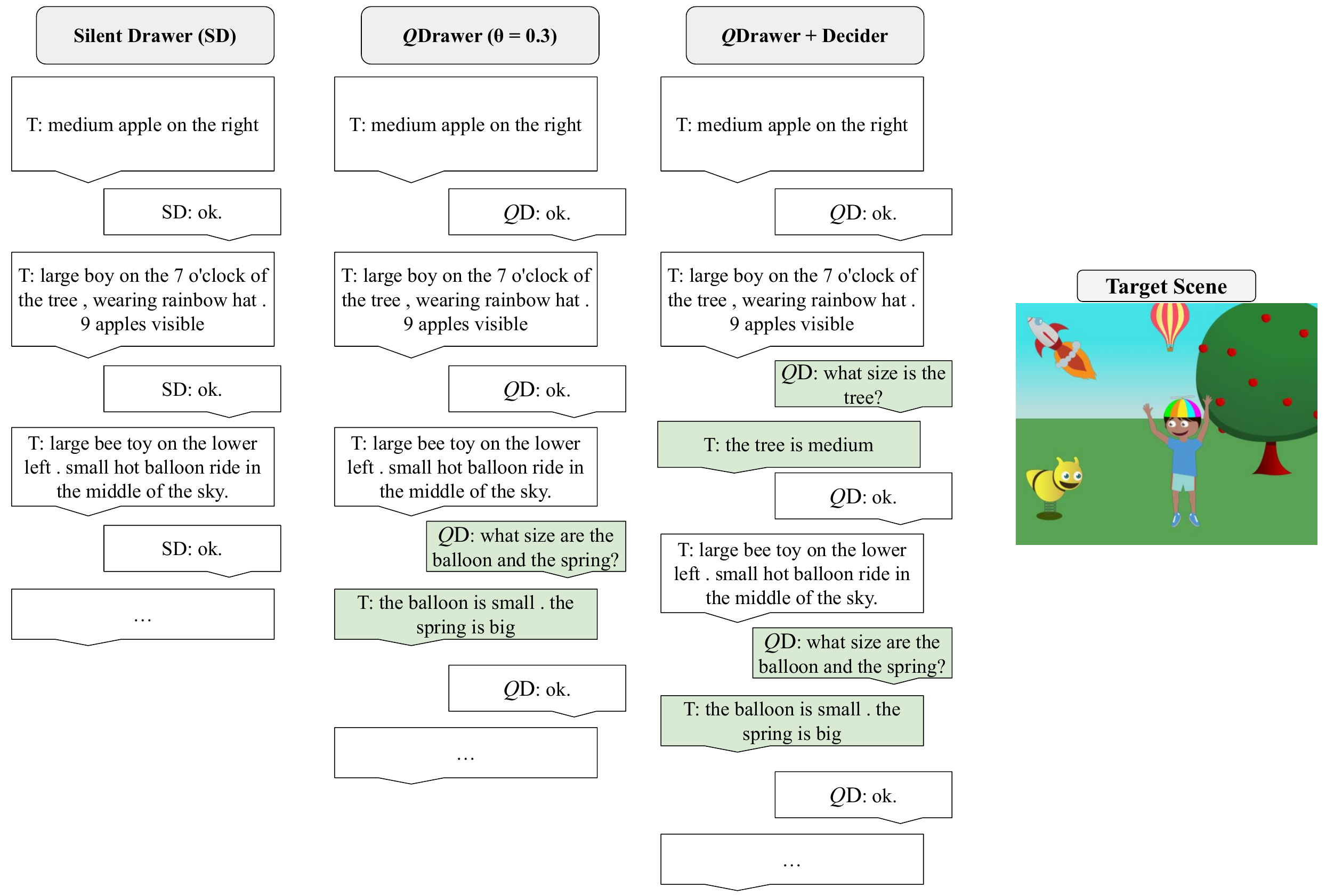} }
  \caption{Examples of dialogues and clarification questions asked by different models. \textit{T} stands for Teller, \textit{SD} for Silent Drawer, and \textit{QD} for \emph{Q}Drawer. Clarification exchanges are highlighted in green.}
  \label{fig:dialogue_examples}
\end{figure*}

Figure \ref{fig:dialogue_examples} shows some examples of clarification questions asked by different models for the same dialogue.

\section{The CoDraw Dataset and Models}
\label{app:codraw-data-models}
The Codraw dataset \citep{kim-etal-2019-codraw} is available at this URL: \url{https://github.com/facebookresearch/CoDraw}. CoDraw is licensed under Creative Commons Attribution-NonCommercial 4.0 International Public License. The CoDraw dataset is fully anonymized.
The Silent drawer model is available at this address: \url{https://github.com/facebookresearch/codraw-models}. This repository is licensed under Creative Commons Attribution-NonCommercial 4.0 International Public License. We use the released data and models consistently with their intended use. Codraw-iCR (v1/v2) is licensed under Creative Commons Attribution-NonCommercial 4.0 International Public License. The CoDraw dataset contains 9993 dialogues, with a total number of 138K utterances. The dataset is randomly split into training, validation, and test set with the following proportions: 80\%, 10\%, 10\%, respectively.

\section{\emph{Q}Drawer Details}
\label{appendix:QDrawer}
The \emph{Q}Drawer model has the same architecture as the \textit{Silent} Drawer described in Section \ref{sec:silent}. The model has 3.347.390 parameters. For training configuration, we follow the original paper \citep{kim-etal-2019-codraw}. We trained the model on a NVIDIA RTX A5000 GPU for 15 epochs ($\approx$ 1 minute/epoch). In \emph{Q}Drawer, we compute the entropy using the SciPy library (\url{https://docs.scipy.org/doc/scipy/index.html}). As mentioned in Section {\ref{sec:asking_clarification_questions}}, we slightly modify the loss function of the \textit{Silent} Drawer. More specifically, the \textit{Silent} Drawer is trained using a combination of losses, including a cross-entropy loss for categorical decisions on which clip art pieces to add to the canvas at a given dialogue turn. This loss considers which new cliparts human players added to the scene at each turn of the dialogue. If the Drawer player updates some attributes of a clipart that was added to the canvas in previous turns (for instance, changing its size or location), this clipart is not included in the loss function. This step is crucial because, at inference time, only the cliparts whose score exceeds a threshold are added to and updated in the canvas. To address this issue, at each dialogue turn, we include in the loss function of \emph{Q}Drawer not only the new cliparts added to the canvas but also the ones already in the canvas that are modified by the Drawer at that turn. In this way, clarification exchanges on cliparts added in previous dialogue turns can be properly addressed.

\begin{table*}[!ht]
\resizebox{\textwidth}{!}{
\begin{tabular}{clcccc}
\multicolumn{6}{c}{\textit{Average Precision}}                                                                                                                                                                                       \\ \cline{3-6} 
\multicolumn{1}{l}{}                                 & \multicolumn{1}{l|}{}           & \multicolumn{4}{c|}{\textbf{traning epoch}}                                                                                                 \\ \cline{3-6} 
\multicolumn{1}{l}{}                                 & \multicolumn{1}{l|}{}           & \multicolumn{1}{c|}{\textbf{1}}  & \multicolumn{1}{c|}{\textbf{5}}  & \multicolumn{1}{c|}{\textbf{10}}  & \multicolumn{1}{c|}{\textbf{15}}  \\ \hline
\multicolumn{1}{|c|}{\multirow{5}{*}{\rotatebox[origin=c]{90}{\textbf{seed}}}} & \multicolumn{1}{l|}{\textbf{0}} & \multicolumn{1}{c|}{.176 (.134)} & \multicolumn{1}{c|}{.186 (.150)} & \multicolumn{1}{c|}{.181 (.143)}  & \multicolumn{1}{c|}{.205 (.143)}  \\ \cline{2-6} 
\multicolumn{1}{|c|}{}                               & \multicolumn{1}{l|}{\textbf{1}} & \multicolumn{1}{c|}{.179 (.150)} & \multicolumn{1}{c|}{.171 (.152)} & \multicolumn{1}{c|}{.185 (.1.38)} & \multicolumn{1}{c|}{.172 (.126)}  \\ \cline{2-6} 
\multicolumn{1}{|c|}{}                               & \multicolumn{1}{l|}{\textbf{2}} & \multicolumn{1}{c|}{.168 (.142)} & \multicolumn{1}{c|}{.181 (.139)} & \multicolumn{1}{c|}{.176 (.137)}  & \multicolumn{1}{c|}{.196 (.1.48)} \\ \cline{2-6} 
\multicolumn{1}{|c|}{}                               & \multicolumn{1}{l|}{\textbf{3}} & \multicolumn{1}{c|}{.194 (.145)} & \multicolumn{1}{c|}{.190 (.153)} & \multicolumn{1}{c|}{.213 (.154)}  & \multicolumn{1}{c|}{.189 (.161)}  \\ \cline{2-6} 
\multicolumn{1}{|c|}{}                               & \multicolumn{1}{l|}{\textbf{4}} & \multicolumn{1}{c|}{.202 (.156)} & \multicolumn{1}{c|}{.179 (.145)} & \multicolumn{1}{c|}{.176 (.125)}  & \multicolumn{1}{c|}{.180 (.135)}  \\ \hline
\end{tabular}

\quad

\begin{tabular}{clcccc}
\multicolumn{6}{c}{\textit{Similarity Score}}                                                                                                                                                                                    \\ \cline{3-6} 
\multicolumn{1}{l}{}                                 & \multicolumn{1}{l|}{}           & \multicolumn{4}{c|}{\textbf{traning epoch}}                                                                                             \\ \cline{3-6} 
\multicolumn{1}{l}{}                                 & \multicolumn{1}{l|}{}           & \multicolumn{1}{c|}{\textbf{1}} & \multicolumn{1}{c|}{\textbf{5}} & \multicolumn{1}{c|}{\textbf{10}} & \multicolumn{1}{c|}{\textbf{15}} \\ \hline
\multicolumn{1}{|c|}{\multirow{5}{*}{\rotatebox[origin=c]{90}{\textbf{seed}}}} & \multicolumn{1}{l|}{\textbf{0}} & \multicolumn{1}{c|}{3.02}       & \multicolumn{1}{c|}{3.33}       & \multicolumn{1}{c|}{3.32}        & \multicolumn{1}{c|}{3.31}        \\ \cline{2-6} 
\multicolumn{1}{|c|}{}                               & \multicolumn{1}{l|}{\textbf{1}} & \multicolumn{1}{c|}{3.05}       & \multicolumn{1}{c|}{3.33}       & \multicolumn{1}{c|}{3.31}        & \multicolumn{1}{c|}{3.34}        \\ \cline{2-6} 
\multicolumn{1}{|c|}{}                               & \multicolumn{1}{l|}{\textbf{2}} & \multicolumn{1}{c|}{3.02}       & \multicolumn{1}{c|}{3.31}       & \multicolumn{1}{c|}{3.34}        & \multicolumn{1}{c|}{3.36}        \\ \cline{2-6} 
\multicolumn{1}{|c|}{}                               & \multicolumn{1}{l|}{\textbf{3}} & \multicolumn{1}{c|}{3.01}       & \multicolumn{1}{c|}{3.30}       & \multicolumn{1}{c|}{3.35}        & \multicolumn{1}{c|}{3.34}        \\ \cline{2-6} 
\multicolumn{1}{|c|}{}                               & \multicolumn{1}{l|}{\textbf{4}} & \multicolumn{1}{c|}{3.04}       & \multicolumn{1}{c|}{3.31}       & \multicolumn{1}{c|}{3.35}        & \multicolumn{1}{c|}{3.35}        \\ \hline
\end{tabular}
}
\caption{On the left, Average Precision of the logistic regression model trained to predict human decisions to ask clarification questions based on model uncertainty. In parenthesis, we report the performance of a random model, which is always significantly lower than its trained counterpart ($p<0.05$). We study the effect of different initialization seeds and training epochs. On the right, we show the corresponding average similarity score reached by the Drawer model at the end of the dialogue in the CoDraw task. }
\label{table:regression_ap_ss}
\end{table*}

\begin{table*}[!ht]
\centering
\resizebox{0.85\textwidth}{!}{
\begin{tabular}{clcccc}
\multicolumn{6}{c}{\textit{F1 Score - Logistic Regression Model}}                                                                                                                                                                \\ \cline{3-6} 
\multicolumn{1}{l}{}                                 & \multicolumn{1}{l|}{}           & \multicolumn{4}{c|}{\textbf{traning epoch}}                                                                                             \\ \cline{3-6} 
\multicolumn{1}{l}{}                                 & \multicolumn{1}{l|}{}           & \multicolumn{1}{c|}{\textbf{1}} & \multicolumn{1}{c|}{\textbf{5}} & \multicolumn{1}{c|}{\textbf{10}} & \multicolumn{1}{c|}{\textbf{15}} \\ \hline
\multicolumn{1}{|c|}{\multirow{5}{*}{\rotatebox[origin=c]{90}{\textbf{seed}}}} & \multicolumn{1}{l|}{\textbf{0}} & \multicolumn{1}{c|}{0.479}      & \multicolumn{1}{c|}{0.482}      & \multicolumn{1}{c|}{0.477}       & \multicolumn{1}{c|}{0.468}       \\ \cline{2-6} 
\multicolumn{1}{|c|}{}                               & \multicolumn{1}{l|}{\textbf{1}} & \multicolumn{1}{c|}{0.493}      & \multicolumn{1}{c|}{0.499}      & \multicolumn{1}{c|}{0.485}       & \multicolumn{1}{c|}{0.463}       \\ \cline{2-6} 
\multicolumn{1}{|c|}{}                               & \multicolumn{1}{l|}{\textbf{2}} & \multicolumn{1}{c|}{0.475}      & \multicolumn{1}{c|}{0.487}      & \multicolumn{1}{c|}{0.472}       & \multicolumn{1}{c|}{0.481}       \\ \cline{2-6} 
\multicolumn{1}{|c|}{}                               & \multicolumn{1}{l|}{\textbf{3}} & \multicolumn{1}{c|}{0.469}      & \multicolumn{1}{c|}{0.481}      & \multicolumn{1}{c|}{0.496}       & \multicolumn{1}{c|}{0.486}       \\ \cline{2-6} 
\multicolumn{1}{|c|}{}                               & \multicolumn{1}{l|}{\textbf{4}} & \multicolumn{1}{c|}{0.483}      & \multicolumn{1}{c|}{0.479}      & \multicolumn{1}{c|}{0.474}       & \multicolumn{1}{c|}{0.474}       \\ \hline
\end{tabular}

\quad

\begin{tabular}{clcccc}
\multicolumn{6}{c}{\textit{F1 Score - Random baseline}}                                                                                                                                                                          \\ \cline{3-6} 
\multicolumn{1}{l}{}                                 & \multicolumn{1}{l|}{}           & \multicolumn{4}{c|}{\textbf{traning epoch}}                                                                                             \\ \cline{3-6} 
\multicolumn{1}{l}{}                                 & \multicolumn{1}{l|}{}           & \multicolumn{1}{c|}{\textbf{1}} & \multicolumn{1}{c|}{\textbf{5}} & \multicolumn{1}{c|}{\textbf{10}} & \multicolumn{1}{c|}{\textbf{15}} \\ \hline
\multicolumn{1}{|c|}{\multirow{5}{*}{\rotatebox[origin=c]{90}{\textbf{seed}}}} & \multicolumn{1}{l|}{\textbf{0}} & \multicolumn{1}{c|}{0.469}      & \multicolumn{1}{c|}{0.492}      & \multicolumn{1}{c|}{0.509}       & \multicolumn{1}{c|}{0.513}       \\ \cline{2-6} 
\multicolumn{1}{|c|}{}                               & \multicolumn{1}{l|}{\textbf{1}} & \multicolumn{1}{c|}{0.500}      & \multicolumn{1}{c|}{0.525}      & \multicolumn{1}{c|}{0.490}       & \multicolumn{1}{c|}{0.488}       \\ \cline{2-6} 
\multicolumn{1}{|c|}{}                               & \multicolumn{1}{l|}{\textbf{2}} & \multicolumn{1}{c|}{0.505}      & \multicolumn{1}{c|}{0.495}      & \multicolumn{1}{c|}{0.504}       & \multicolumn{1}{c|}{0.503}       \\ \cline{2-6} 
\multicolumn{1}{|c|}{}                               & \multicolumn{1}{l|}{\textbf{3}} & \multicolumn{1}{c|}{0.481}      & \multicolumn{1}{c|}{0.507}      & \multicolumn{1}{c|}{0.474}       & \multicolumn{1}{c|}{0.520}       \\ \cline{2-6} 
\multicolumn{1}{|c|}{}                               & \multicolumn{1}{l|}{\textbf{4}} & \multicolumn{1}{c|}{0.493}      & \multicolumn{1}{c|}{0.514}      & \multicolumn{1}{c|}{0.490}       & \multicolumn{1}{c|}{0.519}       \\ \hline
\end{tabular}
}
\caption{F1 score of the logistic regression model (left) and the performance of a random baseline (right).}
\label{tab:f1}
\end{table*}

\begin{table*}[]
\centering
\resizebox{\textwidth}{!}{
\begin{tabular}{|
c |
c |
c |
c |
c |
c |}
\hline
\textbf{seed} & \textbf{epoch} & \textbf{uncertainty(clipart selection)} & \textbf{uncertainty(size)} & \textbf{uncertainty(orientation)} & \textbf{uncertainty(position)} \\  \hline
0             & 1              & -0.397                        & 0.002                   & -0.041                 & -0.096                             \\ \hline
0             & 5              & -0.392                        & 0.006                   & -0.038                 & -0.012                             \\ \hline
0             & 10             & -0.390                        & 0.004                   & -0.044                 & -0.075                             \\ \hline
0             & 15             & -0.390                        & 0.006                   & -0.044                 & 0.040                              \\ \hline
1             & 1              & -0.397                        & 0.002                   & -0.041                 & -0.087                             \\ \hline
1             & 5              & -0.392                        & 0.006                   & -0.038                 & 0.016                              \\ \hline
1             & 10             & -0.390                        & 0.004                   & -0.044                 & -0.027                             \\ \hline
1             & 15             & -0.390                        & 0.006                   & -0.044                 & 0.073                              \\ \hline
2             & 1              & -0.397                        & 0.002                   & -0.041                 & -0.070                             \\ \hline
2             & 5              & -0.392                        & 0.006                   & -0.038                 & 0.025                              \\ \hline
2             & 10             & -0.390                        & 0.004                   & -0.044                 & -0.054                             \\ \hline
2             & 15             & -0.390                        & 0.006                   & -0.044                 & 0.103                              \\ \hline
3             & 1              & -0.397                        & 0.002                   & -0.041                 & -0.054                             \\ \hline
3             & 5              & -0.392                        & 0.006                   & -0.038                 & -0.037                             \\ \hline
3             & 10             & -0.390                        & 0.004                   & -0.044                 & 0.007                              \\ \hline
3             & 15             & -0.390                        & 0.006                   & -0.044                 & -0.015                             \\ \hline
4             & 1              & -0.397                        & 0.002                   & -0.041                 & -0.167                             \\ \hline
4             & 5              & -0.392                        & 0.006                   & -0.038                 & -0.019                             \\ \hline
4             & 10             & -0.390                        & 0.004                   & -0.044                 & -0.048                             \\ \hline
4             & 15             & -0.390                        & 0.006                   & -0.044                 & -0.017                             \\ \hline
\end{tabular}
}
\caption{Correlation coefficients of the logistic regression model for each uncertainty type extracted from the Drawer model. We show the coefficients for each model configuration, as defined by its initialization seed and number of training epochs.}
\label{tab:coeff}
\end{table*}

\begin{table*}[!ht]
\centering
\resizebox{\textwidth}{!}{
\begin{tabular}{|c|c|c|c|c|c|c|c|}
\hline
\textbf{}          & \textbf{\# utts.}      & \textbf{Clip.}                    & \textbf{Size}      & \textbf{Loc.}                     & \textbf{Examples}                                                                                                                                  & \textbf{\% CQ}          & \textbf{CQ type} \\ \hline
\multirow{2}{*}{A} & \multirow{2}{*}{3805} & \multirow{2}{*}{1}                & \multirow{2}{*}{1} & \multirow{2}{*}{$\geq 1$} & \textit{\begin{tabular}[c]{@{}c@{}}- large pine tree on right half of top \\ and right edge cut off\end{tabular}}                                  & \multirow{2}{*}{7.6\%}  & other 48.3\%     \\
                   &                       &                                   &                    &                                   & \textit{\begin{tabular}[c]{@{}c@{}}- middle scene swing, small size. facing right. \\ red part above horizon\end{tabular}}                         &                         & loc. 43.8\%      \\ \hline
\multirow{2}{*}{B} & \multirow{2}{*}{240}  & \multirow{2}{*}{$\geq 1$} & \multirow{2}{*}{-} & \multirow{2}{*}{0}                & \textit{\begin{tabular}[c]{@{}c@{}}- little girl running towards her friend that has \\ a pie in his hand\end{tabular}}                            & \multirow{2}{*}{48.8\%} & loc. 48.3\%      \\
                   &                       &                                   &                    &                                   & \textit{- girl and boy, boy is on swing set}                                                                                                       &                         & other 35.6\%     \\ \hline
\multirow{2}{*}{C} & \multirow{2}{*}{2274} & \multirow{2}{*}{$\geq 1$} & \multirow{2}{*}{0} & \multirow{2}{*}{-}                & \textit{- a tree to the left a few inches from the edge}                                                                                           & \multirow{2}{*}{23.1\%} & other 40.6\%     \\
                   &                       &                                   &                    &                                   & \textit{\begin{tabular}[c]{@{}c@{}}- left side is a bear, facing right, left arm cut \\ from scene. bear cut in half by horizon line\end{tabular}} &                         & size 27.1\%      \\ \hline
\multirow{2}{*}{D} & \multirow{2}{*}{76}   & \multirow{2}{*}{1}                & \multirow{2}{*}{-} & \multirow{2}{*}{0}                & \textit{- it 's at a park with a side and a sandbox}                                                                                               & \multirow{2}{*}{46.1\%} & loc. 57.1\%      \\
                   &                       &                                   &                    &                                   & \textit{- up in the tree are 4 balloons . small size}                                                                                              &                         & other 22.9\%     \\ \hline
\multirow{2}{*}{E} & \multirow{2}{*}{1388} & \multirow{2}{*}{1}                & \multirow{2}{*}{0} & \multirow{2}{*}{-}                & \textit{- boy far lower right , facing left with happy face}                                                                                       & \multirow{2}{*}{19.5\%} & size 40.0\%      \\
                   &                       &                                   &                    &                                   & \textit{\begin{tabular}[c]{@{}c@{}}- girl sitting on ground , fingers touching edge \\ of left screen , 1 2 '' from bottom screen\end{tabular}}    &                         & other 30.4\%     \\ \hline
\multirow{2}{*}{F} & \multirow{2}{*}{5328} & \multirow{2}{*}{1}                & \multirow{2}{*}{-} & \multirow{2}{*}{-}                & \textit{\begin{tabular}[c]{@{}c@{}}- small maple tree on left . almost all leaves \\ above grass line\end{tabular}}                                         & \multirow{2}{*}{10.1\%} & other 40.1\%     \\
                   &                       &                                   &                    &                                   & \textit{- left top small sun left edge little bit cut}                                                                                                      &                         & loc. 34.8\%      \\ \hline
\end{tabular}}
\caption{Cluster analysis by grouping together similar instructions. \textit{\# utts.} refers to the number of utterances in the cluster; \textit{Clip.}, \textit{Size}, and \textit{Loc.} refer to the occurrences of clipart, size, and location keywords, respectively. \textit{\% CQ} refers to the percentage of clarification questions following the given set of instructions. ``-'' means no restrictions on that attribute. 
}
\label{tab:clusters}
\end{table*}

\end{document}